\documentclass[runningheads]{llncs}
\usepackage{graphicx}
\usepackage{verbatim}
\usepackage{xcolor}
\usepackage[linesnumbered,ruled,vlined]{algorithm2e}

\SetCommentSty{mycommfont}

\begin{document}
\title{ProtoAL: Interpretable Deep Active Learning with prototypes for medical imaging}
\titlerunning{ProtoAL: Interpretable Deep Active Learning}
%
\author{Iury B. de A. Santos\inst{1}\orcidID{0000-0001-7234-6877} \and
André C.P.L.F. de Carvalho\inst{1}\orcidID{0000-0002-4765-6459}}
\authorrunning{Santos and Carvalho}
%
\institute{Instituto de Ciências Matemáticas e de Computação (ICMC), University of São Paulo (USP), São Carlos, Brazil \\
\email{iuryandrade@usp.br}, \email{andre@icmc.usp.br}}
\maketitle              
\begin{abstract}
The adoption of Deep Learning algorithms in the medical imaging field is a prominent area of research, with high potential for advancing AI-based Computer-aided diagnosis (AI-CAD) solutions. However, current solutions face challenges due to a lack of interpretability features and high data demands, prompting recent efforts to address these issues. In this study, we propose the ProtoAL method, where we integrate an interpretable DL model into the Deep Active Learning (DAL) framework. This approach aims to address both challenges by focusing on the medical imaging context and utilizing an inherently interpretable model based on prototypes. We evaluated ProtoAL on the Messidor dataset, achieving an area under the precision-recall curve of 0.79 while utilizing only 76.54\% of the available labeled data. These capabilities can enhances the practical usability of a DL model in the medical field, providing a means of trust calibration in domain experts and a suitable solution for learning in the data scarcity context often found.

\end{abstract}
\section{Introduction}
Machine learning (ML) is a field of artificial intelligence that has been rapidly growing in recent years. In particular, the Deep Learning (DL) approach, with large Artificial Neural Network (ANN) architectures, has been extensively explored due to its impressive results in various knowledge fields such as genetics \cite{senior2020improved}, image processing \cite{minaee2021image,jiao2019survey}, and natural language processing \cite{otter2020survey,minaee2021deep}.

In medicine, AI-based Computer-aided diagnosis (AI-CAD) solutions often utilize ML and DL, closely linked to computer vision, to assist in medical diagnoses based on images like Magnetic resonance imaging (MRI), X-rays, Computed tomography (CT) and conventional photographs. Despite considerable community interest, practical application of AI-CAD solutions encounters obstacles, including the lack of interpretability features in models. These models are often perceived as black-box models, making it challenging for humans to understand their internal reasoning, which raises trust issues among experts and regulatory concerns \cite{Litjens_2017,Esteva_2021,Vellido_2020}.

Current AI-CAD solutions often lack robustness, being susceptible to biases during training and failing to provide experts with confidence estimations or limitations regarding the results. This poses challenges, especially in healthcare settings, where less experienced professionals may overly rely on computational models \cite{gaube2021ai,bates2021potential,Litjens_2017}. ProtoPNet, introduced by Chen et al. (2019) \cite{chen2019looks}, is a Deep Neural Network (DNN) architecture aiming to enhance interpretability features within DNN models. During inference, ProtoPNet showcases prototypes that share similar features with the input image in a 'bag-of-features' format.

Several studies have applied the ProtoPNet architecture in medical image analysis. For instance, Jafari et al. (2021) \cite{mohammadjafari2021using} utilized ProtoPNet to classify MRI brain scans as either healthy or indicative of Alzheimer's disease. Vaseli et al. (2023) \cite{vaseli2023protoasnet} introduced ProtoASNet, a modified version of ProtoPNet tailored for handling spatio-temporal data and integrating aleatory uncertainty estimation into prototypes, particularly in the context of aortic stenosis. In their 2024 publication, Wei, Tam, and Tang \cite{wei2024mprotonet} presented the MProtoNet model, an interpretable network similar to ProtoPNet and XProtoNet \cite{kim2021xprotonet}, specifically designed for tumor brain classification tasks in multi-parametric MRI (mpMRI) analysis, featuring modifications enabling training without annotation maps.

The use of DL also faces issues with limited availability of large datasets, especially in supervised learning. Besides data being abundant, labeling requires expert input, resulting in high demands in costs and time. To address this issues, Deep Active Learning (DAL) emerges as a feasible approach, extending Active Learning (AL) \cite{settles2009active} concepts to work with DNN architectures. DAL assumes that a model with comparable results can be achieved using carefully selected training instances, even with less training data, witch is particularly advantageous when dealing with large, unlabeled datasets that would be expensive to label. In DAL, only select instances undergo labeling by experts and are then used to train the model. This interactive process construct a training dataset enriched with more informative and significant instances.

Several works, including \cite{zhao2021dsal,wu2021covid,nath2022warm,belharbi2021deep,di2019deep}, have explored the DAL  approach in medicine, demonstrating promising results in achieving satisfactory outcomes with less data compared with models trained on the entire dataset.  Smailagic et al. (2020) \cite{smailagic2020medal} proposed the O-medal method, which employs online training and eliminates the necessity of the complete model re-training at each DAL cycle. This method presents a more feasible scheme within the context of DNNs. 

There has been limited research on the intersection of DAL and interpretable models. Phillips, Chang, and Friedler (2018) \cite{sahiner2019deep} used LIME interpretability to help experts understand model choices and decide when to accept labeling queries. Das et al. (2019) \cite{das2019active} applied AL to anomaly detection, enriching experts with explanations of the model's decisions using a tree-based ensemble. Liu et al. (2022) \cite{liu2020deep} investigated local interpretations in AL scheme to select and label instances. Similarly, Mondal and Ganguly's (2020) trained a explainer model alongside a classifier model to choose new instances based on the dissimilarity of their explanations compared to already labeled instances.  

This work introduces ProtoAL, a novel method aimed at integrating an interpretable DNN model into the DAL framework, specifically tailored for medical image analysis. ProtoAL utilizes interpretability through prototypes, providing explanations based on image patches from the training dataset, aligning closely with clinical practices in a visually intuitive manner. To assess its effectiveness, ProtoAL was compared against baseline models using robust metrics such as area under the precision-recall curve (AUPRC). Additionally, we examined how the DAL framework impacts the objective of reducing the required training instances to achieve results comparable to models trained conventionally.

\section{Methods}
\begin{figure}
    \centering
    \includegraphics[scale=0.5]{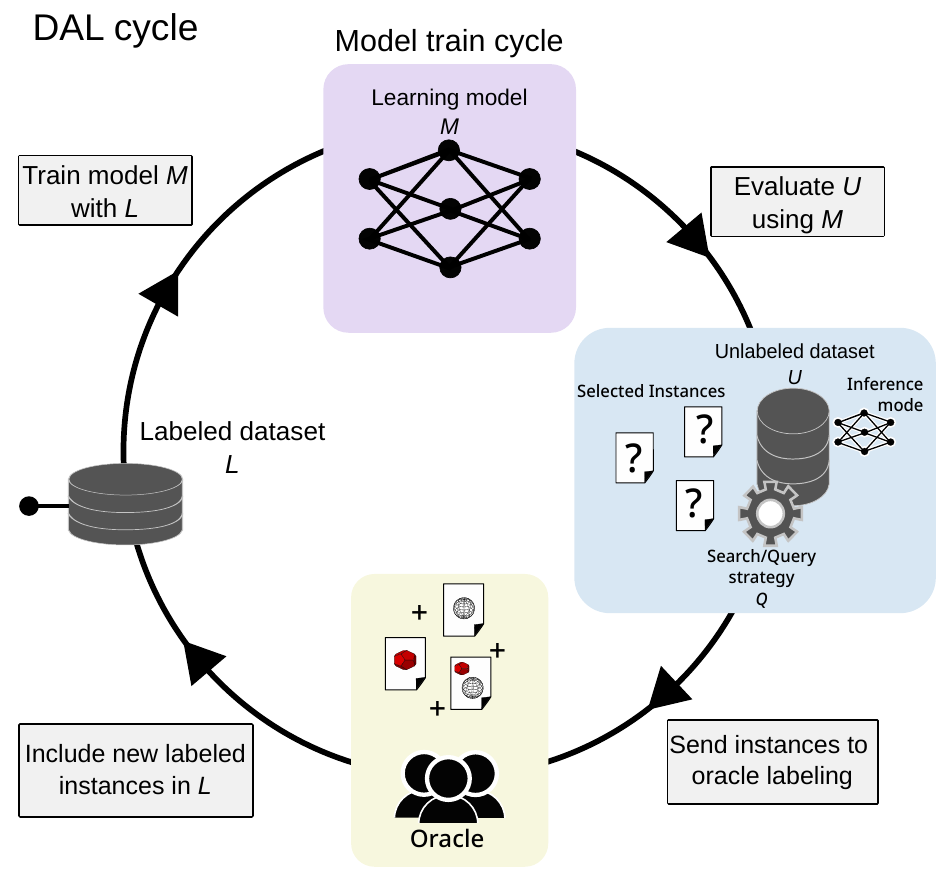}
    \caption{Schematic view illustrating the DAL and model training cycles. In the DAL cycle, labeled instances are added to $\mathcal{L}$ by selecting unlabeled instances from $\mathcal{U}$ using a search strategy. Meanwhile, the learning model M undergoes training iterations within each DAL cycle.}
    \label{fig:protoal}
\end{figure}

Our method, ProtoAL, aims to integrate interpretable DL into the DAL framework, with focus on the AI-CAD to medical imaging. The DAL framework allows train a learning model, $M$, using a training set, $\mathcal{L}$, consisting of selected instances based on a search strategy $Q$. These criteria typically target uncertainty instances or aim to enhance diversity. Initially, these selected instances are part of a large, unlabeled dataset $\mathcal{U}$, and are labeled by an oracle $O$ before being added to $\mathcal{L}$. The DAL framework assumes that selected instances offer pertinent information about the problem, which enables a model achieve comparable performance to one trained on a fully random instance dataset while needing fewer examples. This is particularly beneficial in medical context, where large datasets of unlabeled data exists, but labeled datasets suitable for DL training are scarce. With careful selection of instances, the DAL framework allows reduce the expenses associated with expert labeling. Figure ~\ref{fig:protoal} illustrates our workflow, divided in two cycles: the outer DAL cycle, referring to the process described above; and the training of the interpretable model itself, with its inner workings.

The DAL cycle begins with none or a small number of instances in $\mathcal{L}$, which trains the learning model. At the end of the iteration, the model \textit{M} performs inference in the entire set $\mathcal{U}$. The instances are then evaluated using a search strategy, where $n$ instances are chosen for labeling. After labeling, those instances are included in $\mathcal{L}$. The cycle is repeated until a stop condition is reached. This condition could be a satisfactory performance, budget constraints for labeling new instances, or depletion of the $\mathcal{U}$ dataset. 

The ProtoPNet was the architecture of choice for the model \textit{M}, which performs its own training cycle. The ProtoPNet and its training are detailed in the Subsection ~\ref{protopnet_method}.

\subsection{Deep Active Learning}
We utilize the O-medal method\cite{smailagic2020medal} to define our DAL framework. The O-medal operate similarly to the method described above but differs in two aspects: (\textit{i}) it allows an online approach, eliminating the need to retrain the model from scratch at each DAL iteration; and (\textit{ii}) the search strategy is based on the distance between the feature embedding of the $\mathcal{U}$ instances and $\mathcal{L}$ instances.

The O-medal improves the DNN training workflow by avoiding the need to retrain the model from scratch at each DAL iteration. It offers a more reliable, better fitting, and computationally efficient approach. Also, $\mathcal{L}$ does not include all previously labeled instances. Instead, it consists of newly labeled instances and a partition $p$ of the previously labeled data, which reduces computational costs.

\subsection{Uncertainty estimation with MC Dropout}
Uncertainty estimation is one of the most commonly used estimates for ranking and selection of new instances for labeling. A common approach in the DNN context involves using Monte Carlo Dropout (\textit{MC Dropout})\cite{gal2016dropout}, with the Dropout technique \cite{hinton2012improving,srivastava2014dropout} working as a Bayesian approximation. The Dropout serves to regularize DNNs, reducing the overtrainig by randomly dropping units of the networks during training to prevent co-adapting. With random dropouts, its approximates the effect of combining different neural architectures \cite{hinton2012improving,srivastava2014dropout}.

During inference runs of the $\mathcal{U}$ instances in the \textit{MC Dropout}, the dropout layers remain unfrozen while all other trainable layers are frozen. By performing $T$ forward steps and averaging across each instance of the $\mathcal{U}$ set, the values are ranked, and the $n$ most uncertain instances are labeled and added to the $\mathcal{L}$ set \cite{gal2016dropout}.

\subsection{Prototypical Neural Network} \label{protopnet_method}
The ProtoPNet \cite{chen2019looks} is an interpretability-oriented DNN that utilizes prototypes to explain the reasoning of learning model, making it the preferred model for this study. Unlike explainability in \textit{posthoc} approaches, the ProtoPNet proposes an intrinsically oriented method for achieving interpretable DNN, trough imposed constraints on the learning process, which considers the explanations during training. 

The ProtoPNet model compromises a backbone neural architecture $f$, such as VGGNet \cite{simonyan2014very}, ResNet \cite{he2016deep}, DenseNet \cite{huang2017densely}, or any other selected architecture, extended by a prototypical layer $g_\textbf{p}$ and a fully connected layer $h$. The input feature extraction is realized by $f$, while $g_\textbf{p}$ learns $m$ prototypes \textbf{P} $=\{\textbf{p}_j\}^m_{j=1}$ of $H_{1} \times W_{1} \times D$ shape. The prototype activation pattern acts as a patch representing some prototypical image patch in the original pixel space, where $p$ can be understood as a latent representation of some prototypical part. The squared $L^2$ distances between the input patches and the prototypes of $p$ are inverted into similarity scores, forming an activation map where the values indicate how strongly a prototypical part is present in the image. The activation map is reduced using global max-pooling to a single similarity score, indicating how much a prototypical part is present in some input image patches. The $m$ similarity scores produced by $g_\textbf{p}$ are multiplied by the outputs of the fully connected layer $h$ and normalized using softmax, yielding the predicted probabilities of the image. To ensure that they reflect the training examples, they are projected onto training instances.

The main advantage of ProtoPNet is provide visual explanations of the predictions, which it optimizes and learns alongside the classifier. These prototypes are related to instances from the training set, representing real cases and attributes. This type of interpretability closely aligns with the daily practices of experts in image-based medical applications, eliminating the need for learning new explanatory techniques. 

\section{Experiments}
\subsection{Dataset}
We utilized the Messidor dataset \cite{decenciere2014feedback} of diabetic retinopathy. It compromises 1200 color images of the eyes fundus, obtained from three different ophthalmologic departments. The images comes in varying resolutions: $1440\times960$, $2240\times1488$ or $2304\times1536$ pixels.

Experts evaluated each image and classified it based on the retinopathy grade and risk of macular edema. Retinopathy grade ranges from 0 (normal) to 3, considering the number of microaneurysms, hemorrhages and the presence of neovascularization. Corrections were made by correcting mislabeling and removing duplicated images files according to the instructions in the dataset download page. Following the preprocessing outlined in \cite{smailagic2020medal}, we grouped the retinopathy grades as healthy (DR = 0) or diseased (DR $\geq$ 1). The risk of macular edema feature was not used in the experiments. The images were resized to $512 \times 512$, and data were augmented by randomly applying rotations up to 15 degrees, horizontal flips and scaling in the range [0.9, 1]. The dataset was divided into train, validation and test sets with 759, 190 and 238 instances, respectively. 

\subsection{Baselines}
We compared the ProtoAL method in two perspectives, aiming observe both the interpretability and DAL framework factors. For this, we adopted three baselines, targeting distinct contexts:

\subsubsection{Vanilla ResNet-18 model.} A vanilla ResNet-18 model, trained conventionally and without interpretability features, with access the entire training dataset from beginning. ResNet-18 was adopted due it be the backbone model of the ProtoPNet. This baseline assesses the performance of ResNet-18 trained on the dataset without any modifications explored in this work.

\subsubsection{ProtoPNet standalone baseline.} The aim of this baseline is to evaluate the performance of the interpretability model used in ProtoAL without incorporating it into the DAL framework. Additionally, it aims to verify the impact of the DAL framework and compare the performance difference with respect to ResNet-18.

\subsubsection{ProtoAL with random search query.} The ProtoAL method utilizing as search strategy the selection of random instances from $\mathcal{U}$. This baseline aim to observe the impact of the MC Dropout as search strategy in the performance of the ProtoAL method.

The ResNet-18 and ProtoPNet were pretrained on the ImageNet dataset \cite{russakovsky2015imagenet}, both when used as baselines and backbone in the ProtoAL.

\subsection{Implementation details}
As mentioned earlier, the ProtoAL method follows the structure of the O-medal method, while employing a ProtoPNet as DNN model. We conducted a grid search with 10 seeds, varying hyper-parameters for both the DAL method and ProtoPNet model. Table ~\ref{tab:grid_search} presents the values of the hyper-parameters during the grid search. 

\subsubsection{DAL framework implementation.} We conducted runs using both random (ProtoAL-Random) and MC Dropout (ProtoAL-MC) as search strategy. The stop condition for the DAL cycle was determined as no remaining instances left to be labeled in $\mathcal{U}$. The number of runs was dynamically adjusted based on the labeled instances per DAL iteration. The fixed hyper-parameters follows the \cite{smailagic2020medal} work. The $\mathcal{L}$ set initially consisted of 100 randomly selected instances. The percentage of previously labeled examples forming $\mathcal{L}$ was set to 0.875. 

\subsubsection{ProtoPNet.}
A ResNet-18 was used as the backbone DNN of the ProtoAL, with prototype layers featuring 256 channels. We settled 12 prototypes, with 6 prototypes allocated for each class (healthy and diseased). During training, ProtoPNet undergoes warm-up for 5 epochs during the first DAL iteration. After warm-up, joint training is performed for $e$ epochs (see Epochs per AL iter in Table \ref{tab:grid_search}), succeeded by a projection (push) step and last layer optimization for 15 steps. Excepting for warm-up, this cycle repeats for all DAL iterations.

We utilized the Adam optimizer, with learning rates as outlined in \cite{chen2019looks}. Learning rate was decreased exponentially per epoch. All experiments were run on Nvidia Tesla V100 GPU, with 32GB VRAM. The DNN models were implemented with PyTorch 2.0.1 framework and Python 3.10. 

\begin{table}[hbt]
\centering
\caption{Hyper-parameters explored during the grid search step, both in the DAL cycle and in the training model cycle}
\label{tab:grid_search}
\resizebox{.7\columnwidth}{!}{%
\begin{tabular}{|l|c|}
\hline
\multicolumn{1}{|c|}{Hyper-parameter} & Values                                     \\ \hline
Seeds                                & (0, 1, 2, 5, 10, 12, 42, 123, 1234, 12345) \\ \hline
MC dropout steps                     & (10, 30, 50)                               \\ \hline
Instances to label per DAL iter       & (10 ,20, 30)                               \\ \hline
Batch size                           & (32, 64)                                   \\ \hline
Epochs per DAL iter                   & (5, 10, 20)                                \\ \hline
\end{tabular}%
}
\end{table}

\section{Results}
During grid search, we trained a total of 540 ProtoAL models. We selected the best model configuration using the area under the precision-recall curve (AUPRC) metric, based on validation set results. The best run was achieved by a model trained with a batch size of 32, employing 10 iterations of MC Dropout per instance, and selecting 30 new examples at each DAL iteration. For the joint optimization phase, the ProtoPNet model underwent 10 training epochs. Table \ref{tab:results} presents a comparison of results between ProtoAL and the baselines when evaluated on the test set.

\begin{table}[]
\centering
\caption{Evaluation results of the ProtoAL-MC and baselines in relation to AUPRC, F1-Score, Precision, Recall and Accuracy}
\label{tab:results}
\resizebox{.7\textwidth}{!}{%
\begin{tabular}{|l|ccccc|}
\hline
\multicolumn{1}{|c|}{Model} & AUPRC  & F1-Score & Precision & Recall & Accuracy \\ \hline
ResNet-18                    & 0.8462 & 0.8699   & 0.9067   & 0.8359    & 0.8655   \\ \hline
ProtoPnet                    & 0.7900 & 0.8273   & 0.8512   & 0.8046    & 0.8193   \\ \hline
ProtoAL-Random               & 0.7773 & 0.7999   & 0.8571   & 0.75      & 0.7983   \\ \hline
\textbf{ProtoAL-MC}          & 0.7935 & 0.8181   & 0.8684   & 0.7734    & 0.8151   \\ \hline
\end{tabular}%
}
\end{table}

The ProtoAL-MC achieved an AUPRC of 0.79, along with an F1-Score and accuracy of 0.81. The ProtoAL-Random, utilizing random search query, achieved 0.77 in AUPRC and 0.79 in both F1-Score and accuracy. These results demonstrate that the MC Dropout search strategy yielded superior results, particularly evident in the AUPRC metric.

ProtoPNet and ResNet-18 demonstrated comparable or superior results to ProtoAL in specific metrics. It's worth noting that both baselines had access to the entire training dataset from the start, being more exposed to the instances. However, ProtoPNet exhibits a decline in performance compared to ResNet-18, possibly due to the added complexity of optimization resulting from the inclusion of interpretability features. Consequently, ProtoAL's performance remains consistent and similar to that of the ProtoPNet baseline.

\begin{figure}[hbt]
    \centering
    \includegraphics[scale=.3]{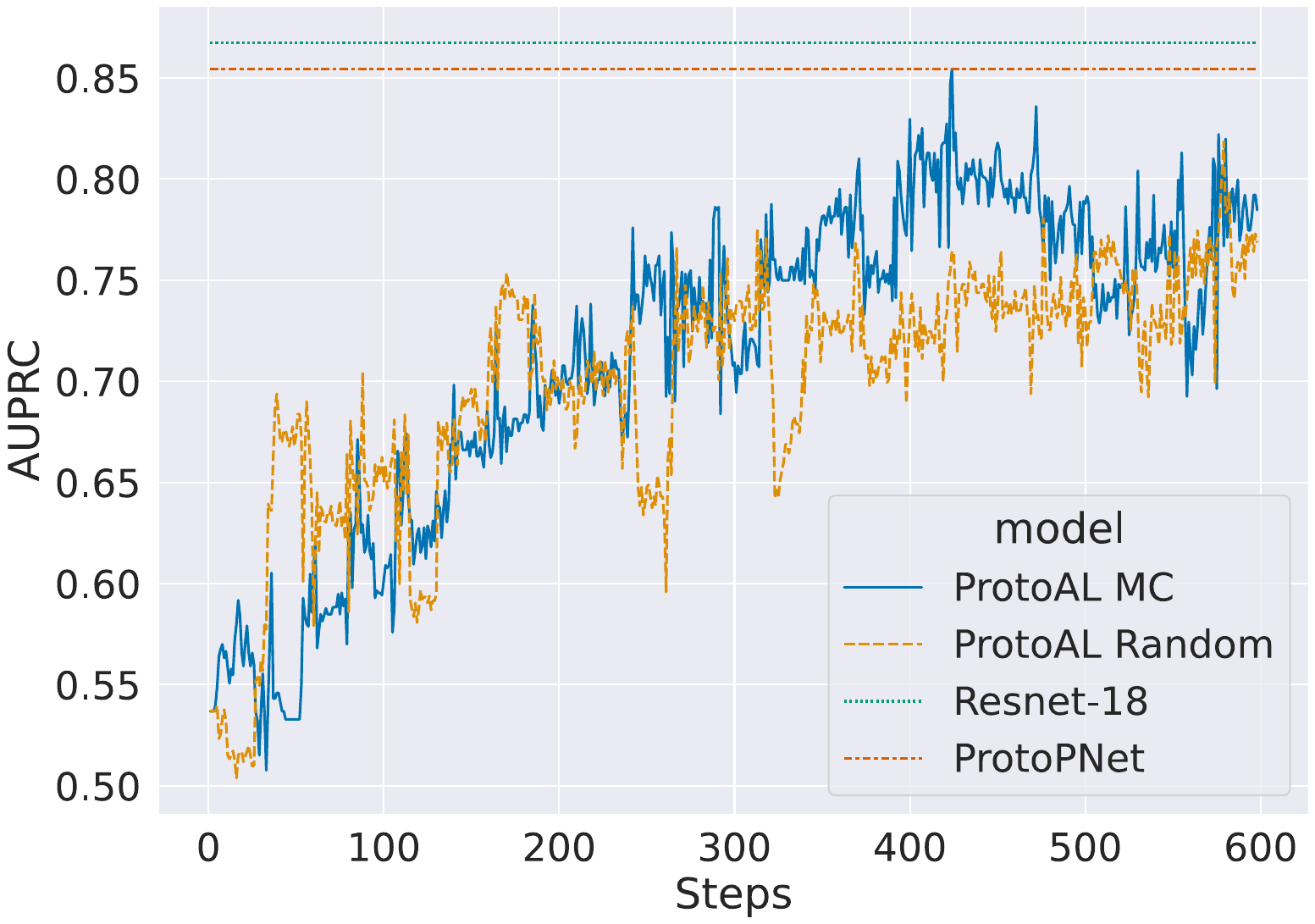}
    \caption{Comparisons of ProtoAL MC method and the baselines, evaluated on the validation set}
    \label{fig:protoal_results}
\end{figure}

The ProtoAL-MC model achieved results comparable to the ProtoPNet baseline in the 17th iteration out of 23 total, with the $\mathcal{L}$ dataset consisting of 512 instances, having labeled 581 instances and leaving 178 instances in the $\mathcal{U}$ set. By this stage, ProtoAL-MC required only 76.54\% of the training instances, out of the initial total of 759 examples, to achieve comparable results to models trained with all available examples. Figure \ref{fig:protoal_results} presents a comparison between ProtoAL-MC and the baseline models across DAL cycle steps when evaluated on the validation set, with ProtoAL-MC achieving its best performance in approximately 400 steps out of a total of 600. The steps are in respect to the total of model training steps (joint, push and last only phases) during all DAL iterations.

\section{Discussion and conclusion}

Our method, ProtoAL, integrates an interpretable DNN model with prototypes into a DAL framework, specifically tailored for the AI-CAD context of medical imaging. Quantitative results presented in Table \ref{tab:results} 
demonstrate the success of providing a interpretable model while utilizing a reduced amount of training data. This addresses two key challenges in the adoption of AI in medical contexts: lack of interpretability and scarcity of labeled datasets. The evaluation primarily focuses on the AUPRC metric, which considers both precision and recall, crucial aspects in the healthcare domain.

ProtoAL offers interpretability features lacking in the ResNet-18 baseline, with a lower requirement for training examples. Despite ProtoAL's seemingly lower performance compared to ResNet-18, these characteristics demonstrate its unique strengths. They enhance the practical usability of ProtoAL as an AI-CAD solution while maintaining a performance level similar to that of the ProtoPNet model, albeit with reduced training data demands. 

Future research could explore enhanced integration of interpretability features within the DAL framework, including leveraging information from prototype components to refine search strategies during DAL cycles. Additionally, explore enhancements in the model to promote prototype diversity and automate the selection of the optimal number of prototypes is warranted.

Integrating an interpretable model into the DAL cycle could enhance the reliability of AI-CAD solutions in practical medical contexts. This integration exploits the capacity to comprehend DNN model decisions and simplifies training in situations with abundant unlabeled data but limited labeled datasets.

\subsubsection*{Acknowledgements.} 
This study was financed in part by the Coordenação de Aperfeiçoamento de Pessoal de Nível Superior – Brasil (CAPES) – Finance Code 001 and grant \#2022/05788-4, São Paulo Research Foundation (FAPESP).

%
%
%
\bibliographystyle{splncs04}
\bibliography{references}

\begin{thebibliography}{10}
\providecommand{\url}[1]{\texttt{#1}}
\providecommand{\urlprefix}{URL }
\providecommand{\doi}[1]{https://doi.org/#1}

\bibitem{bates2021potential}
Bates, D.W., Levine, D., Syrowatka, A., Kuznetsova, M., Craig, K.J.T., Rui, A.,
  Jackson, G.P., Rhee, K.: The potential of artificial intelligence to improve
  patient safety: a scoping review. NPJ digital medicine  \textbf{4}(1), ~1--8
  (2021)

\bibitem{belharbi2021deep}
Belharbi, S., Ben~Ayed, I., McCaffrey, L., Granger, E.: Deep active learning
  for joint classification \& segmentation with weak annotator. In: Proceedings
  of the IEEE/CVF Winter Conference on Applications of Computer Vision. pp.
  3338--3347 (2021)

\bibitem{chen2019looks}
Chen, C., Li, O., Tao, D., Barnett, A., Rudin, C., Su, J.K.: This looks like
  that: deep learning for interpretable image recognition. Advances in neural
  information processing systems  \textbf{32} (2019)

\bibitem{das2019active}
Das, S., Islam, M.R., Jayakodi, N.K., Doppa, J.R.: Active anomaly detection via
  ensembles: Insights, algorithms, and interpretability. arXiv preprint
  arXiv:1901.08930  (2019)

\bibitem{decenciere2014feedback}
Decenci{\`e}re, E., Zhang, X., Cazuguel, G., Lay, B., Cochener, B., Trone, C.,
  Gain, P., Ordonez, R., Massin, P., Erginay, A., et~al.: Feedback on a
  publicly distributed image database: the messidor database. Image Analysis \&
  Stereology  \textbf{33}(3),  231--234 (2014)

\bibitem{di2019deep}
Di~Scandalea, M.L., Perone, C.S., Boudreau, M., Cohen-Adad, J.: Deep active
  learning for axon-myelin segmentation on histology data. arXiv preprint
  arXiv:1907.05143  (2019)

\bibitem{Esteva_2021}
Esteva, A., Chou, K., Yeung, S., Naik, N., Madani, A., Mottaghi, A., Liu, Y.,
  Topol, E., Dean, J., Socher, R.: Deep learning-enabled medical computer
  vision. npj Digital Medicine  \textbf{4}(1), ~5 (Dec 2021).
  \doi{10.1038/s41746-020-00376-2}

\bibitem{gal2016dropout}
Gal, Y., Ghahramani, Z.: Dropout as a bayesian approximation: Representing
  model uncertainty in deep learning. In: international conference on machine
  learning. pp. 1050--1059. PMLR (2016)

\bibitem{gaube2021ai}
Gaube, S., Suresh, H., Raue, M., Merritt, A., Berkowitz, S.J., Lermer, E.,
  Coughlin, J.F., Guttag, J.V., Colak, E., Ghassemi, M.: Do as ai say:
  susceptibility in deployment of clinical decision-aids. NPJ digital medicine
  \textbf{4}(1), ~1--8 (2021)

\bibitem{he2016deep}
He, K., Zhang, X., Ren, S., Sun, J.: Deep residual learning for image
  recognition. In: Proceedings of the IEEE conference on computer vision and
  pattern recognition. pp. 770--778 (2016)

\bibitem{hinton2012improving}
Hinton, G.E., Srivastava, N., Krizhevsky, A., Sutskever, I., Salakhutdinov,
  R.R.: Improving neural networks by preventing co-adaptation of feature
  detectors. arXiv preprint arXiv:1207.0580  (2012)

\bibitem{huang2017densely}
Huang, G., Liu, Z., Van Der~Maaten, L., Weinberger, K.Q.: Densely connected
  convolutional networks. In: Proceedings of the IEEE conference on computer
  vision and pattern recognition. pp. 4700--4708 (2017)

\bibitem{jiao2019survey}
Jiao, L., Zhang, F., Liu, F., Yang, S., Li, L., Feng, Z., Qu, R.: A survey of
  deep learning-based object detection. IEEE access  \textbf{7},
  128837--128868 (2019)

\bibitem{kim2021xprotonet}
Kim, E., Kim, S., Seo, M., Yoon, S.: Xprotonet: diagnosis in chest radiography
  with global and local explanations. In: Proceedings of the IEEE/CVF
  conference on computer vision and pattern recognition. pp. 15719--15728
  (2021)

\bibitem{Litjens_2017}
Litjens, G., Kooi, T., Bejnordi, B.E., Setio, A.A.A., Ciompi, F., Ghafoorian,
  M., van~der Laak, J.A., van Ginneken, B., Sánchez, C.I.: A survey on deep
  learning in medical image analysis. Medical Image Analysis  \textbf{42},
  60–88 (Dec 2017). \doi{10.1016/j.media.2017.07.005}

\bibitem{liu2020deep}
Liu, Q., Liu, Z., Zhu, X., Xiu, Y.: Deep active learning by model
  interpretability. arXiv preprint arXiv:2007.12100  (2020)

\bibitem{minaee2021image}
Minaee, S., Boykov, Y.Y., Porikli, F., Plaza, A.J., Kehtarnavaz, N.,
  Terzopoulos, D.: Image segmentation using deep learning: A survey. IEEE
  transactions on pattern analysis and machine intelligence  (2021)

\bibitem{minaee2021deep}
Minaee, S., Kalchbrenner, N., Cambria, E., Nikzad, N., Chenaghlu, M., Gao, J.:
  Deep learning--based text classification: a comprehensive review. ACM
  Computing Surveys (CSUR)  \textbf{54}(3),  1--40 (2021)

\bibitem{mohammadjafari2021using}
Mohammadjafari, S., Cevik, M., Thanabalasingam, M., Basar, A.: Using protopnet
  for interpretable alzheimer's disease classification. In: Canadian Conference
  on AI (2021)

\bibitem{nath2022warm}
Nath, V., Yang, D., Roth, H.R., Xu, D.: Warm start active learning with proxy
  labels and selection via semi-supervised fine-tuning. In: International
  Conference on Medical Image Computing and Computer-Assisted Intervention. pp.
  297--308. Springer (2022)

\bibitem{otter2020survey}
Otter, D.W., Medina, J.R., Kalita, J.K.: A survey of the usages of deep
  learning for natural language processing. IEEE transactions on neural
  networks and learning systems  \textbf{32}(2),  604--624 (2020)

\bibitem{russakovsky2015imagenet}
Russakovsky, O., Deng, J., Su, H., Krause, J., Satheesh, S., Ma, S., Huang, Z.,
  Karpathy, A., Khosla, A., Bernstein, M., et~al.: Imagenet large scale visual
  recognition challenge. International journal of computer vision
  \textbf{115}(3),  211--252 (2015)

\bibitem{sahiner2019deep}
Sahiner, B., Pezeshk, A., Hadjiiski, L.M., Wang, X., Drukker, K., Cha, K.H.,
  Summers, R.M., Giger, M.L.: Deep learning in medical imaging and radiation
  therapy. Medical physics  \textbf{46}(1),  e1--e36 (2019)

\bibitem{senior2020improved}
Senior, A.W., Evans, R., Jumper, J., Kirkpatrick, J., Sifre, L., Green, T.,
  Qin, C., {\v{Z}}{\'\i}dek, A., Nelson, A.W., Bridgland, A., et~al.: Improved
  protein structure prediction using potentials from deep learning. Nature
  \textbf{577}(7792),  706--710 (2020)

\bibitem{settles2009active}
Settles, B.: Active learning literature survey  (2009)

\bibitem{simonyan2014very}
Simonyan, K., Zisserman, A.: Very deep convolutional networks for large-scale
  image recognition. arXiv preprint arXiv:1409.1556  (2014)

\bibitem{smailagic2020medal}
Smailagic, A., Costa, P., Gaudio, A., Khandelwal, K., Mirshekari, M., Fagert,
  J., Walawalkar, D., Xu, S., Galdran, A., Zhang, P., et~al.: O-medal: Online
  active deep learning for medical image analysis. Wiley Interdisciplinary
  Reviews: Data Mining and Knowledge Discovery  \textbf{10}(4),  e1353 (2020)

\bibitem{srivastava2014dropout}
Srivastava, N., Hinton, G., Krizhevsky, A., Sutskever, I., Salakhutdinov, R.:
  Dropout: a simple way to prevent neural networks from overfitting. The
  journal of machine learning research  \textbf{15}(1),  1929--1958 (2014)

\bibitem{vaseli2023protoasnet}
Vaseli, H., Gu, A.N., Ahmadi~Amiri, S.N., Tsang, M.Y., Fung, A., Kondori, N.,
  Saadat, A., Abolmaesumi, P., Tsang, T.S.: Protoasnet: Dynamic prototypes for
  inherently interpretable and uncertainty-aware aortic stenosis classification
  in echocardiography. In: International Conference on Medical Image Computing
  and Computer-Assisted Intervention. pp. 368--378. Springer (2023)

\bibitem{Vellido_2020}
Vellido, A.: The importance of interpretability and visualization in machine
  learning for applications in medicine and health care. Neural Computing and
  Applications  \textbf{32}(24),  18069–18083 (Dec 2020).
  \doi{10.1007/s00521-019-04051-w}

\bibitem{wei2024mprotonet}
Wei, Y., Tam, R., Tang, X.: Mprotonet: A case-based interpretable model for
  brain tumor classification with 3d multi-parametric magnetic resonance
  imaging. In: Medical Imaging with Deep Learning. pp. 1798--1812. PMLR (2024)

\bibitem{wu2021covid}
Wu, X., Chen, C., Zhong, M., Wang, J., Shi, J.: Covid-al: The diagnosis of
  covid-19 with deep active learning. Medical Image Analysis  \textbf{68},
  101913 (2021)

\bibitem{zhao2021dsal}
Zhao, Z., Zeng, Z., Xu, K., Chen, C., Guan, C.: Dsal: Deeply supervised active
  learning from strong and weak labelers for biomedical image segmentation.
  IEEE journal of biomedical and health informatics  \textbf{25}(10),
  3744--3751 (2021)

\end{thebibliography}

\end{document}